# UAV Autonomous Localization using Macro-Features Matching with a CAD Model

**Akkas Haque [1], Ahmed Elsaharti [1], Tarek Elderini [2], Mohamed Atef Elsaharty [1,*] and Jeremiah Neubert [1]**

[1] Department of Mechanical Engineering, University of North Dakota (UND), Upson II Room 266, 243 Centennial Drive, Stop 8359, Grand Forks, ND 58202, USA; akkasuddin.haque@und.edu (A.H.); ahmed.elsaharti@und.edu (A.E); jeremiah.neubert@und.edu (J.N.)

[2] School of Electrical Engineering and Computer Science, University of North Dakota (UND), Upson II Room 369. 243 Centennial Drive, Stop 7165, Grand Forks, ND 58202, USA; tarek.elderini@und.edu

* Correspondence: mohamed.elsaharty@und.edu; Tel.: +1-701-777-5996

**Abstract:** Research in the field of autonomous Unmanned Aerial Vehicles (UAVs) has significantly advanced in recent years, mainly due to their relevance in a large variety of commercial, industrial, and military applications. However, UAV navigation in GPS-denied environments continues to be a challenging problem that has been tackled in recent research through sensor-based approaches. This paper presents a novel offline, portable, real-time in-door UAV localization technique that relies on macro-feature detection and matching. The proposed system leverages the support of machine learning, traditional computer vision techniques, and pre-existing knowledge of the environment. The main contribution of this work is the real-time creation of a macro-feature description vector from the UAV captured images which are simultaneously matched with an offline pre-existing vector from a Computer-Aided Design (CAD) model. This results in a quick UAV localization within the CAD model. The effectiveness and accuracy of the proposed system were evaluated through simulations and experimental prototype implementation. Final results reveal the algorithm's low computational burden as well as its ease of deployment in GPS-denied environments.

## 1. Introduction

Due to the proliferation of Unmanned Aerial Vehicles (UAV) applications in the past decade, vision-based localization of UAVs has been an important and active field of research for several years. This is primarily due to the increase in applications where Global Positioning Systems (GPS) and Global Navigation Satellite System (GNSS) are infeasible [1,2]. Furthermore, auxiliary functions such as object recognition can be integrated into the vision system reducing overall complexity. In indoor applications, GPS signal can be weak, inaccurate or unavailable, thus preventing autonomous UAVs from localizing within buildings which gives rise to multiple complications that hamper navigation efforts [2,3].

Non-vision-based localization in GPS-denied environments either rely on wireless sensor network [4,5], Time of Flight, and Received Signal Strength Indicator (RSS) [6]. Such solutions require special building-wide setup consisting of sensors and/or transmitters to be constructed adequate functionality [5]. Furthermore, the increased cost as the setup footprint increases as well as synchronization between different nodes yields a major complexity in this type of setups [6].

As a resolution to such complications, vision-based localization approaches in GPS-denied environments have been widely discussed in literature. Detection and mapping to the corresponding location in the environment plays a key role in vision-based localization. Most of the approaches discussed in literature rely upon Simultaneous Localization and Mapping (SLAM) systems which create a 3D map of an unknown environment [7]. This helps in determining the relative location of the UAV at any instant within a GPS-denied environment. However, SLAM based systems create a heavy computational burden which yields a bottleneck for a real-time processing scenario [3].

To overcome such a challenge, captured image feature extraction and comparison with a preset model has been discussed in [8–10]. In [8], a localization method is proposed though extraction of lines from the environment using an omnidirectional camera. These lines are then matched with 3D line segments in a CAD model of the environment using a robust matching algorithm. Although this approach brings in numerous innovative features, it does require the robot to be moving only in Special Euclidean group (SE(2)). However, moving in SE(3) would exponentially increase the search space for the initial localization of the robot. Not to mention its need for a relatively bulky omnidirectional camera which could be problematic in some cases. This method has been utilized in [9] as well using the onboard RGB camera to determine the pose of the camera with respect to the environment. This in turn provides complementary drift update to the localization and mapping SLAM derived pose estimation. However, this has been implemented solely for camera pose estimation and never for localization of the UAV. A different approach in [10,11] proposes a method of unknown indoor environment exploration that utilizes the semi-dense nature of the map obtained from Line Segment Detector-SLAM (LSD-SLAM) to generate an octo-map. LSD-SLAM utilizes all the available image information such as lines and edges which leads to a robust and high accuracy denser 3D reconstruction of the environment. This is then used to locate spaces that a UAV can navigate while populating the unknown regions of the space around it. Although such approaches can be effective, the computational burden of these methodologies creates a bottleneck for future added features to the system.

This paper proposes a technique that starts by extracting macro-features [12] from stereo images of the environment (captured during flight) using a Convolutional Neural Network (CNN) [13]. These macro-features are then registered to a SLAM map that is generated during flight. Finally, the registered information is then used to find correspondences between the SLAM map and a 3D CAD model of the environment. This would enable the UAV to quickly and effectively create a transformation between the SLAM map (in-which it is localized by default) and the CAD model, thus localizing the UAV within the CAD model of the environment.

The main contribution of this paper is the construction of a novel feature vector made up of spatial relations between macro-features. This feature vector is to be leveraged to efficiently search for and correlate similarities in macro-features between the SLAM map and the 3D CAD model of the environment. With such a methodology, localization can be achieved in a computationally effective manner as compared to previously discussed approaches. Furthermore, the performance of the proposed system is evaluated via experimental implementation on a UAV that autonomously localizes itself within a building given a 3D CAD model of the building. After localization, the system is further tested by having the UAV autonomously navigate to a set goal position within the CAD model to evaluate the overall accuracy of the proposed localization method.

This paper is divided into five sections. Section 2 explains the utilized methodology from pre-processing to macro-features detection, extraction and matching. The hardware implementation used to test and evaluate the proposed method is discussed in Section 3. Section 4 reveals the results of the experiment and analyzed in terms of detection accuracy, feature vector effectiveness as well as localization accuracy. Finally, conclusions are presented in Section 5.

## 2. Methodology

The proposed localization technique will be utilized in an indoor environment where the macro-features of the environment are doors and windows. The preliminary function is to detect such features using a CNN discussed in section 2.2. For each detected macro-feature, a unique binary description vector is created by encoding the distances and angles between the neighboring macro-features. These description vectors are compared (in section 2.3) to the preprocessed vectors (in section 2.1) in a 3D CAD model. Once multiple matches are found, the UAV can localize itself within the CAD model and therefore would be able to follow a calculated trajectory to the goal position.

### *2.1. Pre-Processing*

Prior to the UAV flight, a Building Information Model (BIM) [14] is loaded using Open Asset Import Library (Assimp) [15] which provides a unified method of accessing the data within. A BIM is a 3D CAD model typically produced during the planning phase of a construction project and includes information about macro-features as well as their locations as shown in Figure 1.

Macro-features are fixed features in the indoor environment such as doors, windows, and vent panels in which the proximity of each feature to the other can be used to identify a particular location on a 3D map. The macro-features from the CAD model are extracted and their centroids' locations are stored in a K-Dimensional Tree (k-d Tree) [16,17] of dimension 3. This facilitates retrieval of information about the nearest neighboring macro-feature of any given reference macro-feature which is required while generating the description vector within the CAD model.

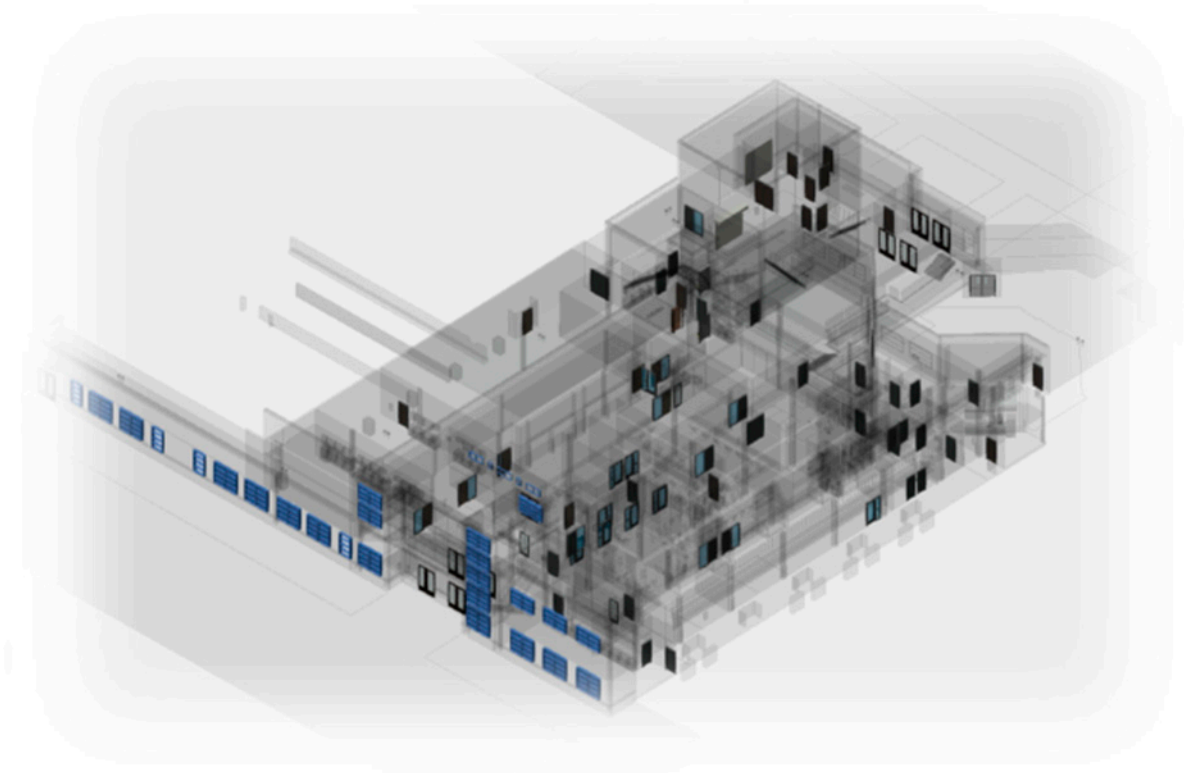

**Figure 1.** The Building Information Model: The required macro-features (doors and windows) are extracted and highlighted. Doors are depicted in black and windows in blue. All the other components are represented in translucent gray.

Using the k-d Tree, vectors are computed for each macro-feature within the CAD model using an algorithm explained in Section 2.3. These are later used to detect correspondences between the detected macro-features during the UAV flight and those present within the CAD model.

### *2.2. Macro-Feature Detection and Extraction Method*

The objective here is to device a method to detect doors and windows (macro-features) reliably and efficiently from observed data while simultaneously registering them to the generated SLAM map. A CNN YOLOv2 [13] that has been previously trained to detect macro-features would ensure a high level of accuracy while maintaining a quick computational speed. Since windows are harder to detect using traditional image processing techniques due to reflections and the presence of objects behind them. YOLOv2 performs real-time detection by applying a single neural network to the entire image. This network divides the image into regions and predicts bounding boxes and probabilities of predictions for each region.

#### 2.2.1. Data Preparation and Training

The training data is collected using short video sequences containing doors and windows from the real datasets. The data is collected from a variety of different sources including the actual indoor flight environment. Care is taken to collect images in different lighting conditions as well as from different distances and angles to ensure training effectiveness. The images are annotated using LabelImg [18] tool. Doors and windows in each image were marked using rectangular boxes and the coordinates were stored in normalized coordinates by expressing them as a fraction of the length and width of the image. Slight perturbation in angle were introduced to the images during augmentation. This was done by applying a rotation to the input images given by,

$$R_z(\theta) = \begin{bmatrix} \cos\theta & -\sin\theta & 0 \\ \sin\theta & \cos\theta & 0 \\ 0 & 0 & 1 \end{bmatrix}, \quad (1)$$

where $R_z$ is a rotation about an axis perpendicular to $xy$ plane and $\theta$ is the angle of rotation. This angle is kept to a maximum of ±20 degrees in 5-degree increments to simulate the maximum possible roll. Furthermore, inverse warping is applied to get rid of holes in the warped images. These holes were artifacts of forward warping where pixels in the warped images are not painted in. This is due to the unavailability of one-to-one mapping from the source images to the destination images.

A total of 1540 images are collected and a total of 12,320 images are generated through augmentation. A total of 80% of the dataset is randomly selected as the training set and the remaining 20% are to validate the detection accuracy. This is done to compute the prevision-recall curve of the detection which provides information to set the cut-off threshold of the detection confidence. Detections are classified into two classes, one for the doors and the other for windows. In addition, to remove detections that looked like neither, a third class is added. This class is not used in the evaluation but leads to an increase in the neural network accuracy.

#### 2.2.2. Detection Refinement

The YOLOv2 provides rectangular boxes bounding the proposed detection with the image space as shown in Figure 2a. Each of these bounding boxes corresponds to a single detection of either a door or a window. These detections are then translated into the actual macro-features in 3D space. To do this, lines are first detected in the area within each of the bounding boxes using Line Segment Detector (LSD) [19,20]. In practice, even though the algorithm does perform quite well line detection, the output is prone to generate broken lines even after rigorous parameter tuning as in Figure 2b. Further refinement is applied to such an issue by joining vertical side lines within a threshold distance from bounded box provided they meet certain criteria. The LSD outputs lines as point pairs denoting the endpoints of the line segments. These endpoints are used to compute the slopes of line segments. Line segments are joined when such line segments have a slope within a threshold angle as well as close endpoints are within a threshold distance. The threshold for the angle is set to 2 degrees and the threshold distance between close endpoints is set to 10 pixels. The longest lines closest to the two vertical sides of the bounds are then taken to denote the vertical sides of the macro-feature as illustrated in Figure 2c. This is similarly illustrated in Figure 2d and Figure 2e on one and multiple doors, respectively.

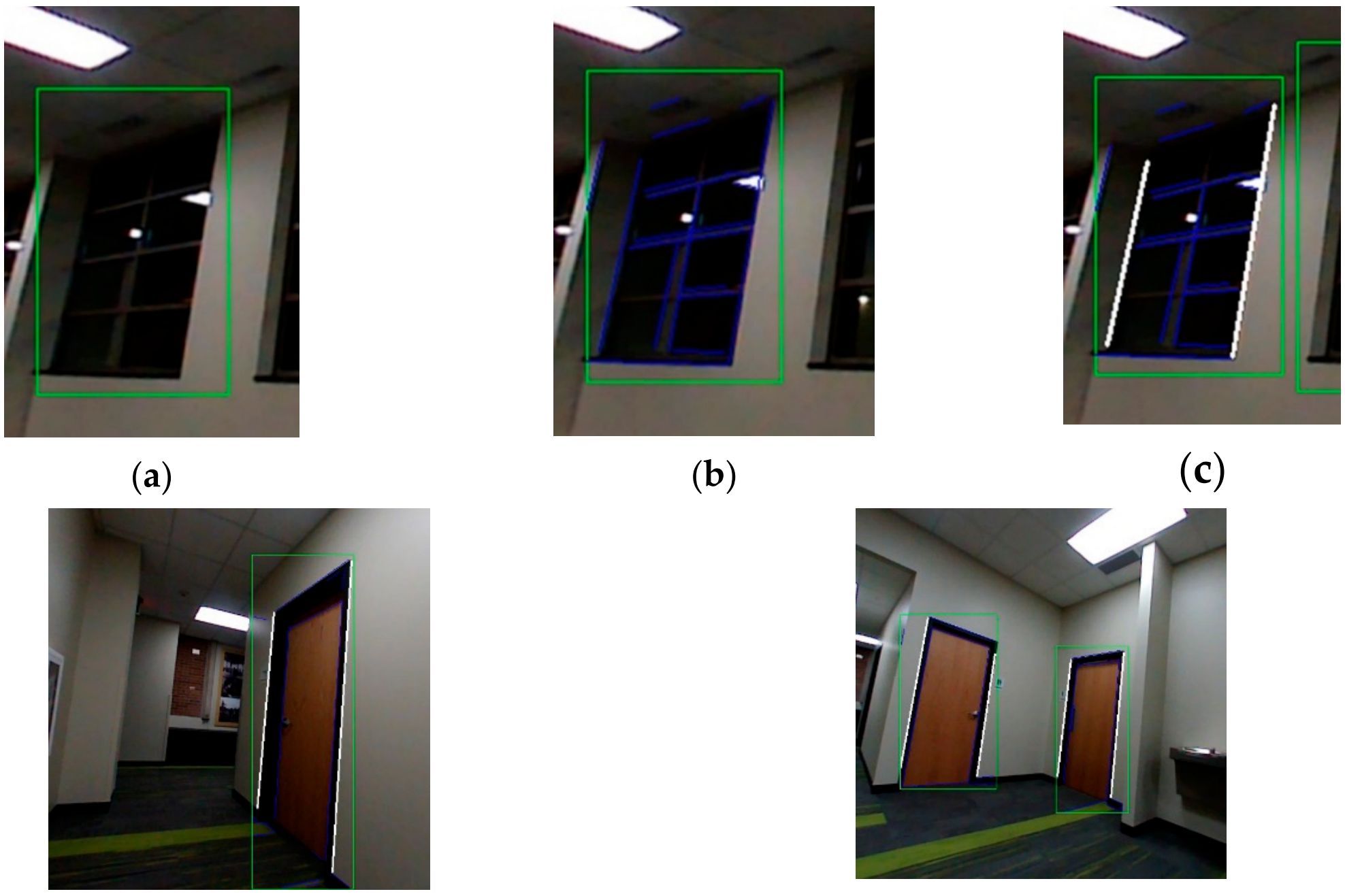

(**a**) (**b**) (**c**) (**d**) (**e**)

**Figure 2.** Macro-feature detection and refinement process from (**a**) YOLOv2 bounding box, (**b**) lines detection using Line Segment Detector (LSD) algorithm, (**c**) vertical sides of the object detected and marked, (**d**) vertical sides of single door detected and marked and (**e**) multiple doors detected and marked.

### 2.2.3. 3D Reconstruction

Using stereo image pairs captured during flight, the 2D lines detected and denoted as the vertical edges of each macro-feature are projected into 3D space. The image pairs are used to compute the depth from disparity at each point in the image. This computation is highly parallelizable and is done on board the UAV. The sampled points are shown in Figure 3a along the 2D vertical side lines detected in Figure 2c and the depth of each point are projected into the SLAM map in Figure 3b. This is computed by:

$$\begin{bmatrix} P_{3x} \\ P_{3y} \\ P_{3z} \end{bmatrix} = d \begin{bmatrix} (P_{2x} - c_x)/f_x \\ (P_{2y} - c_y)/f_y \\ 1 \end{bmatrix}, \quad (2)$$

where $P_{3x}$, $P_{3y}$, and $P_{3z}$ are the coordinates of the point in 3D with respect to the current pose of the camera in the world coordinate system. While $P_{2x}$ and $P_{2y}$ are the coordinates of the point in the 2D image coordinate system with a computed depth of $d$. Camera coordinates and focal lengths are represented by $c_x$, $c_y$ and $f_x$, $f_y$, respectively.

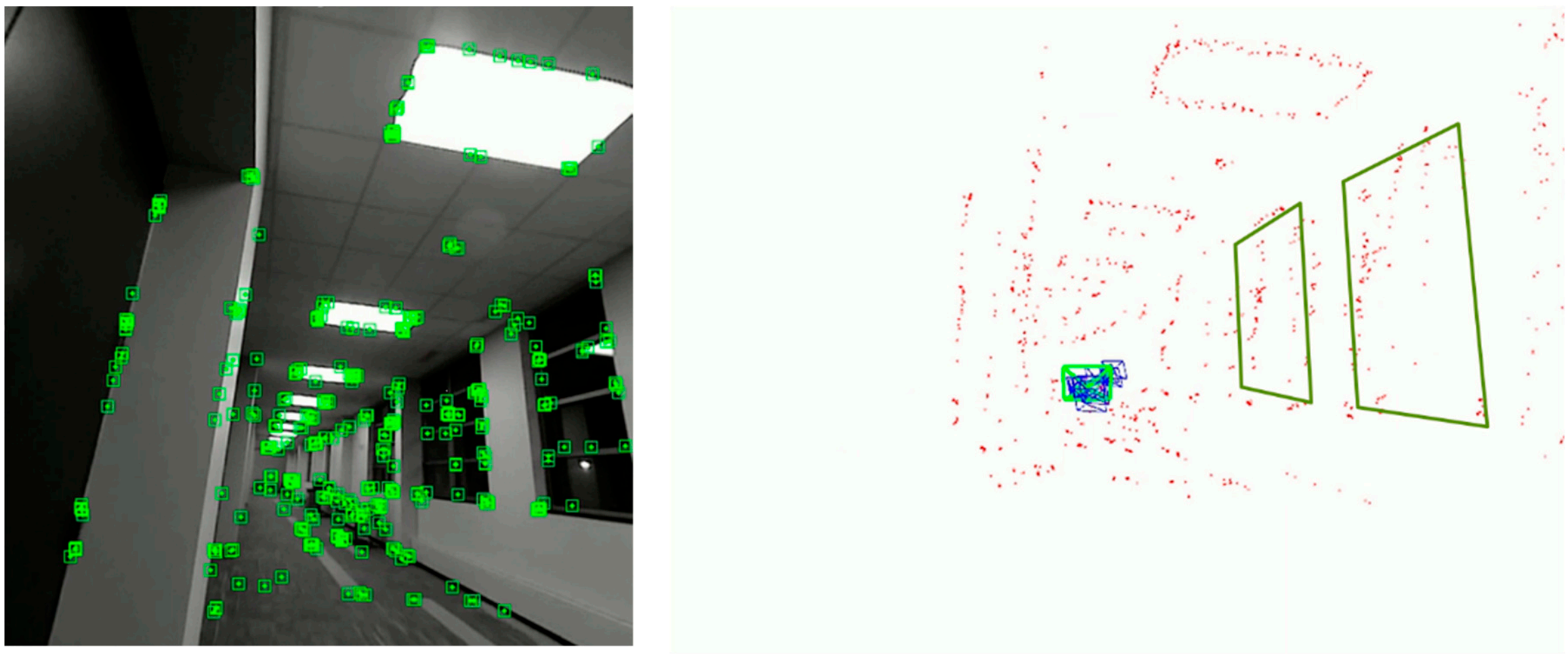

(**a**) (**b**)

**Figure 3.** Reconstruction and registration of macro-features in 3D from (**a**) tracked features to (**b**) the generated Simultaneous Localization and Mapping (SLAM) map showing the 3D reconstructed windows and previously registered keyframes in blue quads.

The sampled points are used to compute the line in 3D using a least-squared line fitting algorithm relying on orthogonal distance. Outliers are then removed by removing points that are at a threshold distance from the computed line. The line is once again recomputed from the inliers using the orthogonal regression used previously. This two-step computation of the line provides a better estimate and is resistant to outliers in practice. The endpoints of the line are computed by finding the projection of the two endpoints of the line segment in 3D onto the fitted line. The four points that make up these line segments now represent the macro-feature in 3D in the view coordinates. Each of these macro-features detected are assigned to the keyframe in which they are first detected during flight.

Keyframes are frames in a graph-based SLAM that are tracked across multiple frames. They consist of the camera pose at an instant along with map points in view of the keyframe and contain the connection to other keyframes in the SLAM map that share closely correlated map points. Note that assigning the detected macro-features to the keyframes ensures spatial consistency in the event of a bundle adjustment step occurring before the localization step is complete.

A homogenous point in 3D is defined by:

$$\mathbf{P} = \begin{bmatrix} P_x \\ P_y \\ P_z \\ 1 \end{bmatrix}, \quad (3)$$

where $P_x$, $P_y$, and $P_z$ are the three scalar coordinates. A transformation, $T \in SE(3)$ is represented by:

$$T = \begin{bmatrix} R & t \\ 0 & 1 \end{bmatrix}, \quad (4)$$

where $R \in SO(3)$ is a rotation matrix and $t$ is a translation.

At any point in time, the locations of the macro-feature points in the world coordinates can be computed by,

$$\mathbf{P}_w = T_w^c \mathbf{P}_c, \quad (5)$$

where $\mathbf{P}_w$ are the homogeneous coordinates of the point in the world coordinate system, $\mathbf{P}_c$ are the homogeneous coordinates of the point in the view coordinate system, and $T_w^c$ is the transformation between the camera and the world coordinate systems.

For the course registration phase that takes place after this step, each of the macro-features computed above are further reduced to their respective centroids in the world coordinate system. The description vector for these centroids are computed according to the following subsection.

*2.3. Macro-Feature Registeration and Matching*

In the proposed system, the UAV is already localized within the SLAM coordinate system. This means that the successful alignment of the CAD model with the SLAM map would result in localization of the UAV within the CAD model which is the main goal. This can be achieved by successfully matching the macro-features between both the CAD model and the SLAM map.

A SLAM algorithm that utilizes stereo images ensures that the SLAM map generated has an accurate estimation of scale. Since the two sets of data have the same scale, a rigid transformation is sufficient to align both sets of data. To compute this transformation, point-to-point correspondences is required between both datasets. Once the correspondences are found, a Random Sample Consensus (RANSAC) [21] based approach uses subsets of these correspondences to determine the best set of corresponding macro-features to utilize and the required transformation is computed.

To speed up the process of finding corresponding macro-features, a novel method is being devised that encodes each macro-feature into a binary macro-feature description vector that is orientation and position invariant. This is the main contribution of the paper and is explained in detail in the following subsections. It is to be noted that the use of a binary description vector ensures fast macro-feature matching due to the utilization of Hamming distance as the distance measurement unit. This is because the computation of a Hamming distances can be done in a single XOR operation between the two binary strings/vectors. The orientation and position invariance help in removing the dependence of a vector on the coordinate system.

To obtain matches between the macro-features in both datasets, the information within the vector needs to be decoupled from both coordinate systems and should mainly be dependent on the macro-features themselves. This is done by encoding the distances and angles between the closest 5 macro-features into the description vector making the data within the vector dependent on only the macro-features and the co-relations between them. Since the SLAM coordinate system has close to accurate scale, the relative distances and orientations between the closest macro-features stay the same in both coordinate systems. Figure 4 illustrates when the macro-features are reduced to their respective centroids, the relative distances and orientations between the macro-features stay the same.

Figure 4a shows portion of the CAD model with a macro-feature whose vector is being calculated while Figure 4b shows the SLAM map generated with the detected macro-features and corresponding labels in both images. The description vector is computed by first retrieving the 5 macro-features closest to a macro-feature for which the vector is being computed. This is done efficiently by utilizing the k-d Tree to store the centroids of the detected macro-features. Next, using the macro-feature closest to the macro-feature for which the computation is occurring as a base, the angles to each of the 4 other macro-features are computed using the formula,

$$\alpha = \arccos\left(\frac{\mathbf{a}.\mathbf{b}}{\|\mathbf{a}\|.\|\mathbf{b}\|}\right). \tag{6}$$

The vector is clarified in Figure 5, where $\mathbf{a}$ is the vector between the current macro-feature and the closest (base) macro-feature, $\mathbf{b}$ is the vector between the current macro-feature and the macro-feature whose angle is to be calculated, and $\boldsymbol{\alpha}$ is the angle between the two vectors.

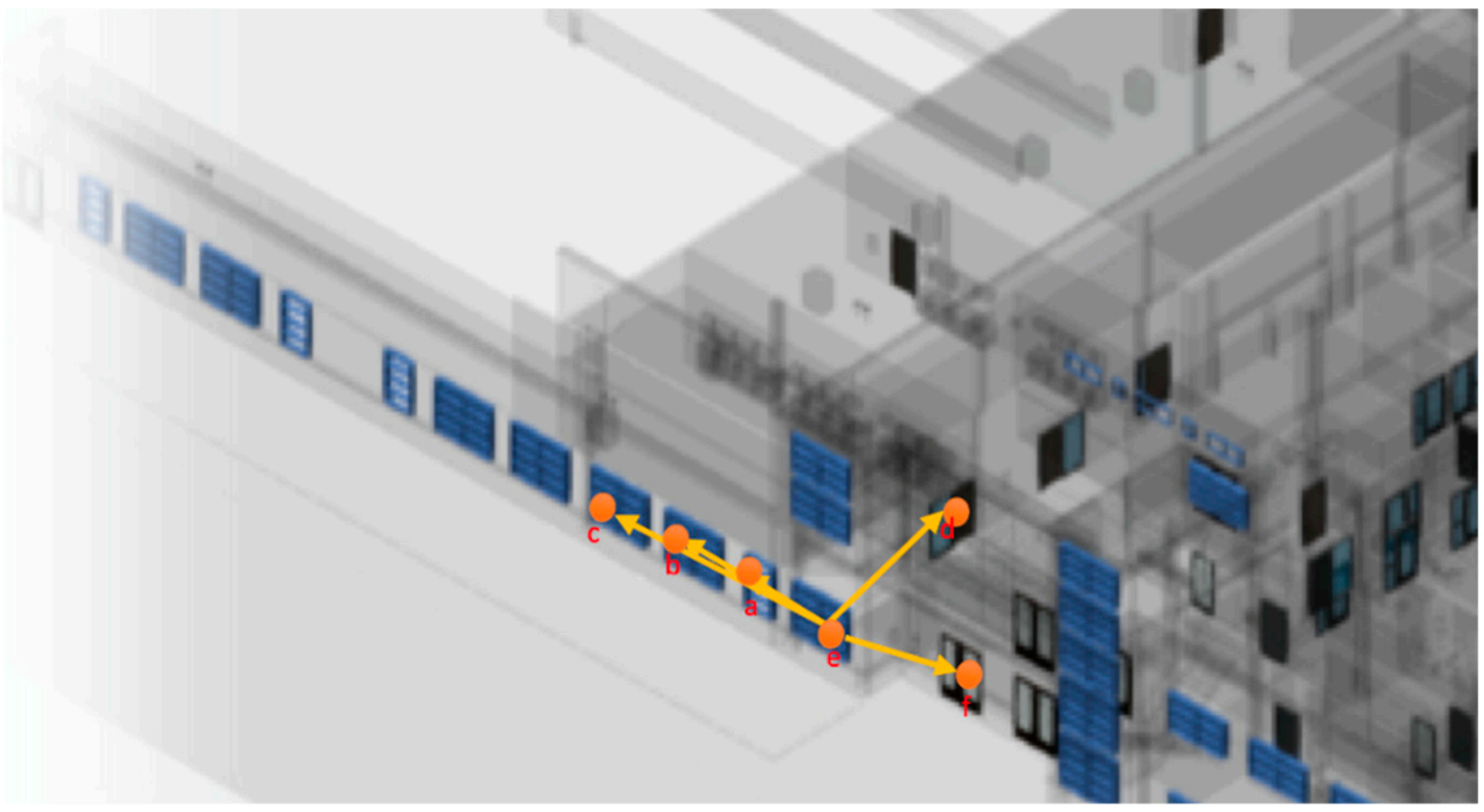

(**a**)

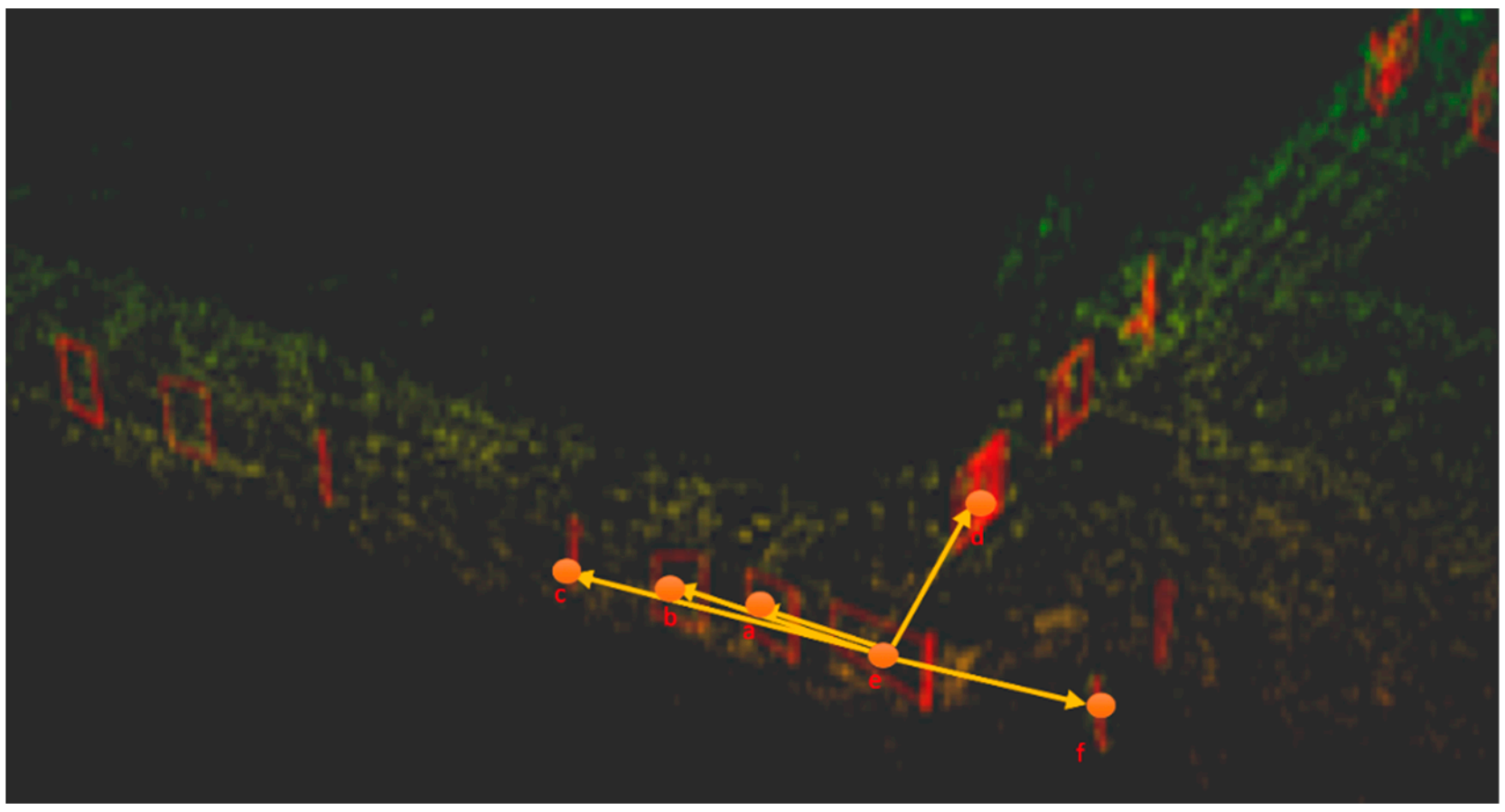

(**b**)

**Figure 4.** Orientation invariance of the macro-feature description vector. The images show the relative distances and orientation of a few selected macro-features, with 'e' being the microfeature of which the description vector is being calculated. Image (**a**) shows the macro-features in the 3D CAD model while image (**b**) shows the same macro-features in the SLAM map.

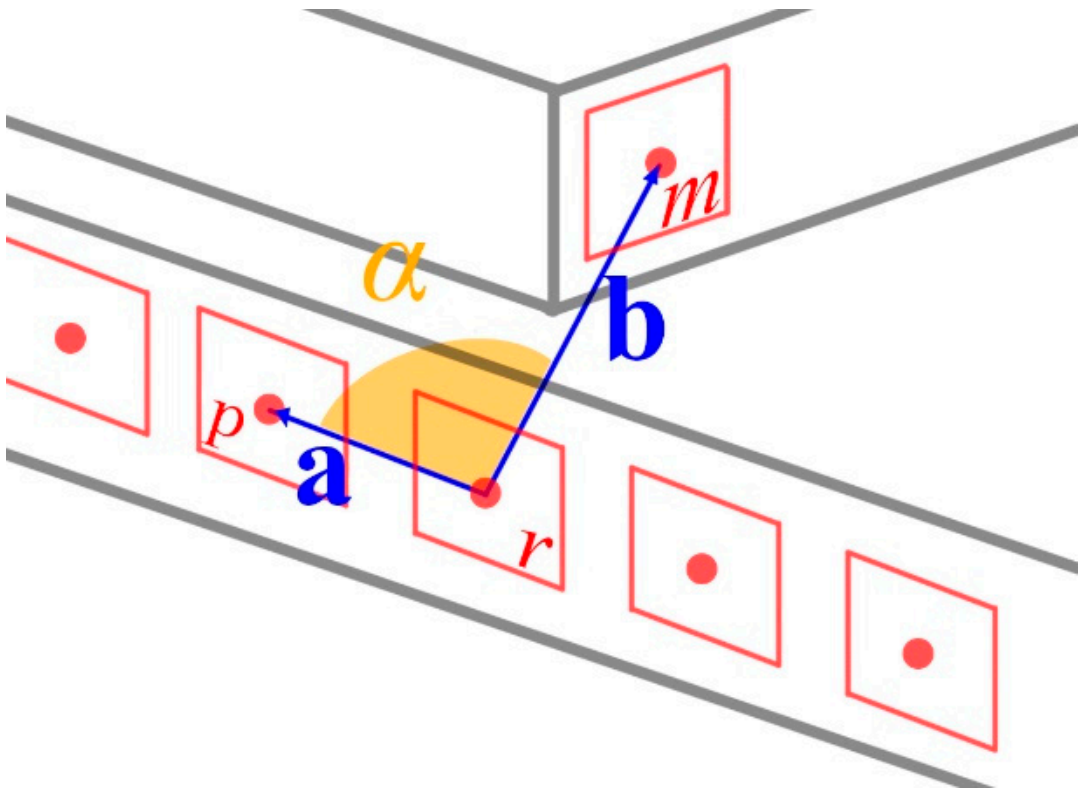

**Figure 5.** Angle calculation—In this figure, $r$ is the macro-feature for which the description vector is being calculated, $p$ is the closest macro-feature to $r$ (taken as a base for the calculation), $m$ is the macro-feature for which the angle relative to $r$ is being calculated, and $\alpha$ is the angle between vectors **a** and **b** that connects $r$, $p$, and $m$.

The encoding of the relative orientations and distances to each of the macro-features is achieved through a lookup table. The lookup table ensures that the values of distances and angles that are close to local means are assigned the same value. This binning process enables the encoding of the distances and the angles into a compact form that can then be used to form the binary descriptive vector. This process is outlined in Algorithm 1. In addition to the distance and orientation information, the first bit of vector (*feature_type*) is used to denote the type of the macro-feature i.e., door (0) or window (1). This is further explained in the next subsection.

---

**Algorithm 1:** Computation of Feature Descriptor

---

**Result:** Descriptor vector, $d$

Extract closest 5 features from k-d Tree with distances;

$v \leftarrow all_5_points$

$x \leftarrow current_pt$

$vec_0 \leftarrow v_0 - x$

**forall** $i$ *in* [1, 2, 3, 4] **do**

    $vec_i \leftarrow v_i - x$

    $dist_i \leftarrow LUT_{dist}(|vec_i, vec_0|)$

    $\angle i \leftarrow LUT_{angle}(vec_i, vec_0)$

**end**

Descriptor vector, $d \leftarrow [feature_type, [dist_1], [\angle 1], [dist_2], [\angle 2], [dist_3], [\angle 3], [dist_4], [\angle 4]$

---

### 2.3.1. Lookup Table and Vector Structure

The lookup table formed in Algorithm 1 is computed through a binning process. Two lookup tables are created, one for angles and the other for distances. These tables are used to group together values that are close to local means. The boundaries of these bins are computed using a Kernel Density Estimation (KDE) process [22].

First, all the possible values are computed during the pre-processing stage using the CAD model. These values are then sorted and grouped into clusters which is achieved by sorting all the values and then estimating the shape of a probability density function $f$ (using a KDE) that provides a representation of the data. The Kernel Density Estimator, $\widehat{f_h}$ is given by:

$$\widehat{f_h}(x) = \frac{1}{n}\sum_{i=1}^{n} K_h(x - x_i) = \frac{1}{nh}\sum_{i=1}^{n}\left(\frac{x - x_i}{h}\right), \tag{7}$$

where $K$ is the kernel (in this case a Gaussian kernel), $K_h$ is the scaled kernel, $h$ is a smoothing parameter and $x_1$, $x_2$, …. $x_n$ are univariate, independent, and identically distributed samples drawn from the distribution function. The bins required for the lookup table are created by estimating and analyzing the shape of the function then dividing the distribution along its local minima as illustrated in Figure 6.

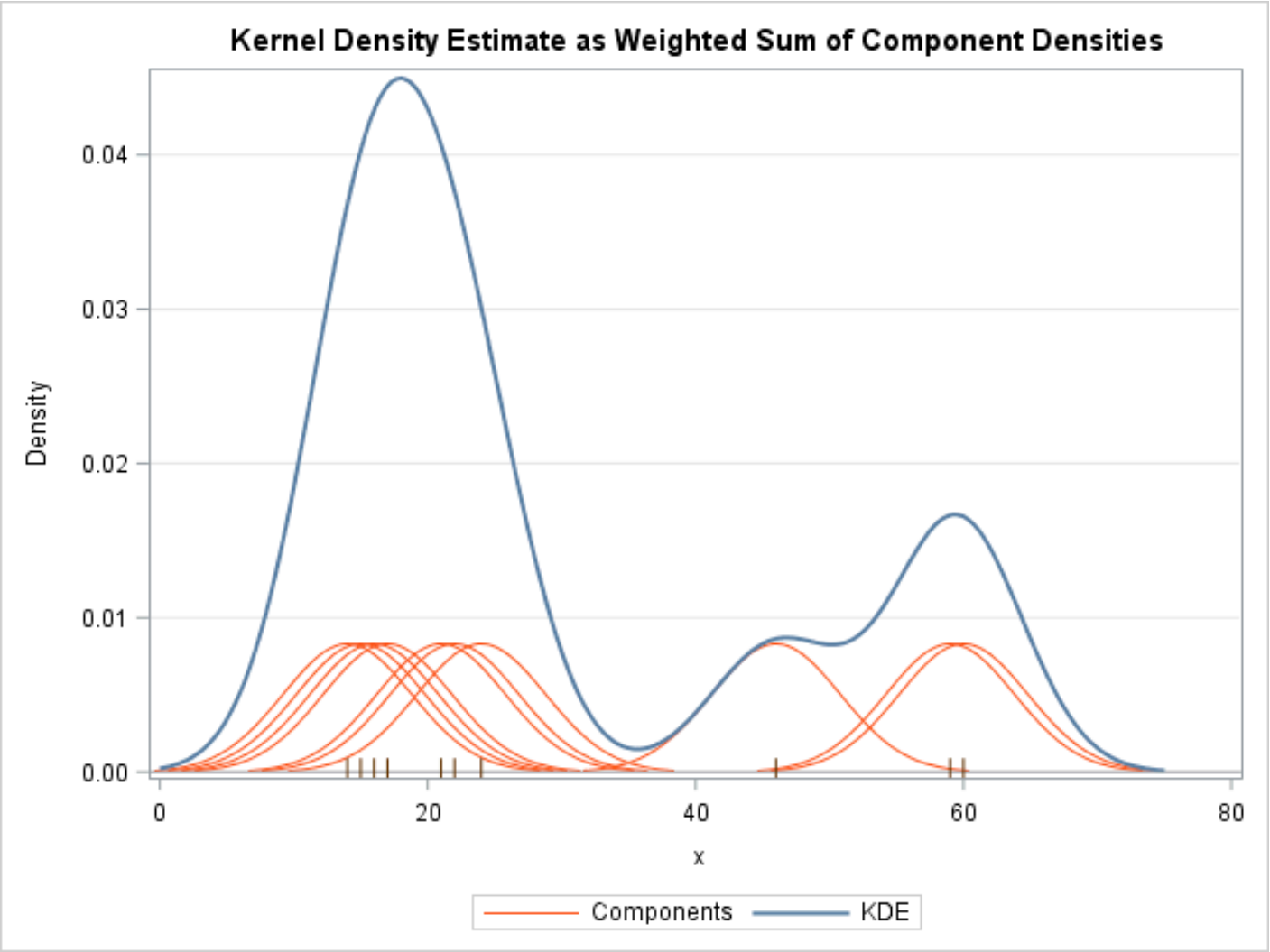

**Figure 6.** The Kernel Density Estimation used to create bins for the lookup table.

By considering the underlying density of this function to be Gaussian and $\hat{\sigma}$ as the standard deviation of values, the value of $h$ can be estimated using Silverman's rule of thumb as follows:

$$h = \left(\frac{4\hat{\sigma}^5}{3n}\right)^{\frac{1}{5}} \approx 1.06\hat{\sigma}n^{-1/5}, \tag{8}$$

Each of the bins are then assigned a number between 0 to the total number of bins, the maximum being 255 to fit an 8-bit width in the binary representation of the number. This way, each neighboring macro-feature gives rise to two 8-bit values, one for the distance and one for the angle. As mentioned earlier that 4 neighboring macro-features are considered to define a specific macro-feature, the total size of the descriptive vector is thus 64 bits. The first bit in the 64-bit vector is usually zero as the number of bits used to represent the number of bins in the angles is less than 8. This enables the use of the first bit to store the type of the macro-feature while using the remaining 7 bits to store the first angle. The final structure of the macro-feature description vector is shown in Figure 7. In this study, the macro-feature type bit takes a 0 or 1 value depending on the feature being a door or window. If the macro-feature types match, the vector score is evaluated using the rest of the bits, if not, the maximum value for the distance is returned i.e., 64.

Since this is a binary vector, matching of the vector is done by finding the Hamming distance between two vectors. The vectors are matched on the basis of the lowest Hamming distance. Hamming distance will compute the number of positions of two equal strings where the corresponding values are different. This is computed very easily and efficiently using the XOR operator and then summing the total number of set bits in the result. Computers with newer hardware usually have support for counting the total number of set bits in CPU instructions. This results in a computational efficient matching mechanism. The complexity for finding matches between the observed macro-features and the CAD macro-features is thus, *O(mn)*, with $m$ being the number of observed macro-features and $n$ being the number of CAD macro-features.

| 1bit | 7bits | 8bits | 8bits | 8bits | 8bits | 8bits | 8bits | 8bits |
|---|---|---|---|---|---|---|---|---|
| Feature Type | angle1 | dist1 | angle2 | dist2 | angle3 | dist3 | angle4 | dist4 |

**Figure 7.** The 64-bit orientation invariant description vector with the first bit used to define the macro-feature type.

### 2.3.2. Initial Registration

During the initial registration phase, macro-features are extracted using the methods outlined in Section 2.2. For each extracted macro-feature, the centroid of the macro-feature is stored in the k-d Tree and the vector is calculated when the macro-feature has more than a threshold number of neighbors. These vectors and the vectors computed from the CAD model during the pre-processing stage are then matches using the methods outlined in Section 2.3.1.

The matching macro-features are then used in a RANSAC which works by selecting the minimum number of parameters needed to estimate a model hypothesis. RANSAC is a robust iterative method that is used to estimate the parameters of a mathematical model from noisy data that has outliers. The generated model is then used to compute the total number of inliers for the model hypothesis. This process is repeated for a total of $N$ times to generate $N$ hypotheses and the hypothesis with the greatest number of inliers is selected as the transformation. The number $N$ is a function of the desired probability of success ($p$). If $\omega$ is the probability of selecting an inlier for each selected point, and $m$ is the number of points required to estimate the model, then $1-\omega^m$ will be the probability of finding at least one outlier among the $m$ points. Meanwhile, $(1-\omega^m)^N$ is thus the probability of never selecting $m$ points that are all inliers. Thus, the total number of hypothesis that need to be generated for a desired probability of success is:

$$N = \log(1-p)\,/\,\log(1-\omega^m). \tag{9}$$

Therefore, $N$ different hypotheses are generated to estimate the model. In this study, the model is a rigid transformation between two 3D point sets. Since this requires at least four point-to-point correspondences, the value of $m$ in this case is four, which are randomly selected from the matched macro-features. The rigid transformation is then computed using a linear least squares method based on Singular Value Decomposition (SVD).

Coplanarity or collinearity of the points give rise to degenerate conditions and must be avoided. Since collinear points are also coplanar it is sufficient to perform the check for coplanarity. The test for coplanarity can be done by computing the scalar triple product, which is given by:

$$|(x_3 - x_1).[(x_2 - x_1) \times (x_4 - x_1)]| < \varepsilon, \tag{10}$$

where $x_1, x_2, x_3, x_4 \in \mathbb{R}^3$are the four distinct points and $\varepsilon = 0.01$ is the margin of error. Any set of points giving rise to a value less than $\varepsilon$ are considered coplanar. If the selected points pass the coplanarity test, they are rejected, and a new set of points are selected for the hypothesis. The hypothesis that generates the greatest number of inliers is then selected as the transformation.

The rigid transformation in SE(3) between the two point sets is computed using the following methodology. If $A := \{\mathbf{a}_i | i = 1, 2, \ldots, n, \mathbf{a}_i \in \mathbb{R}^3\}$ and $B := \{\mathbf{b}_i | i = 1, 2, \ldots, n, \mathbf{b}_i \in \mathbb{R}^3\}$ are two corresponding sets of points, $A$ being the source and $B$ being the destination set, the rigid transformation to be computed is found by minimizing the squared error of the transformed coordinates, which is given by:

$$(R, \mathbf{t}) = \underset{R \epsilon SO(3), \mathbf{t} \in \mathbb{R}^3}{\operatorname{argmin}} \sum_{i=1}^{n} \omega_i \|(R\mathbf{a}_i + \mathbf{t}) - \mathbf{b}_i\|^2, \tag{11}$$

where $R$ is the rotation in *SO(3)* and $\mathbf{t}$ in the translation while $\omega_i > 0$ are the weights assigned to the squared differences for each point pair. The weighted centroids on both sets are computed by:

$$\bar{\mathbf{a}} = \frac{\sum_{i=1}^{n} \omega_i \mathbf{a}_i}{\sum_{i=1}^{n} \omega_i}, \bar{\mathbf{b}} = \frac{\sum_{i=1}^{n} \omega_i \mathbf{b}_i}{\sum_{i=1}^{n} \omega_i}, \tag{12}$$

where $\bar{\mathbf{a}}$ and $\bar{\mathbf{b}}$ are the centroids of the two sets. Vectors are then computed using the point sets and the respective centroids by,

$$x_i := \mathbf{a}_i - \bar{\mathbf{a}}, \qquad y_i := \mathbf{b}_i - \bar{\mathbf{b}}, \qquad i = 1, 2, \ldots\ldots, n \tag{13}$$

where $x_i$ and $y_i$ are the corresponding vectors originating at the respective centroids. The 3 × 3 covariance matrix is then computed using:

$$S = XWY^T \tag{14}$$

where $X$ and $Y$ are the vectors of dimension 3 and $W = \text{diag}(\omega_1, \omega_2, \ldots, \omega_n)$, the weight matrix. The singular value decomposition of

$$S = U\sigma V^T \tag{15}$$

is computed where, $U$ and $V$ are orthogonal unitary matrices, and $\sigma$ is a diagonal matrix with non-negative real numbers along the diagonal which are the singular values of $S$. The rotation matrix $R$ is then given by

$$R = VU^T \tag{16}$$

The translation can then be calculated as

$$\mathbf{t} = \bar{\mathbf{b}} - R\bar{\mathbf{a}} \tag{17}$$

After computing the transformation matrix using the least-squares based RANSAC method outlined above, the transformation is refined by using all the points from all the features. This means that all the four points that make up the corners of the macro-features (doors/windows) in the 3D space are used to compute the least-squares rigid transformation. This gives rise to a better estimation of the transformation.

## 3. Hardware Implementation

The hardware used to implement and test the proposed system comprised of Parrot's SLAM dunk module mounted on top of a Parrot Bebop 2 drone and a ground control station. The SLAM dunk utilizes NVIDIA's Jetson TK1 board along with integrated stereo camera, an Inertial Measurement Unit (IMU), a front facing ultrasonic sensor, a barometer, and a magnetometer as shown in Figure 8.

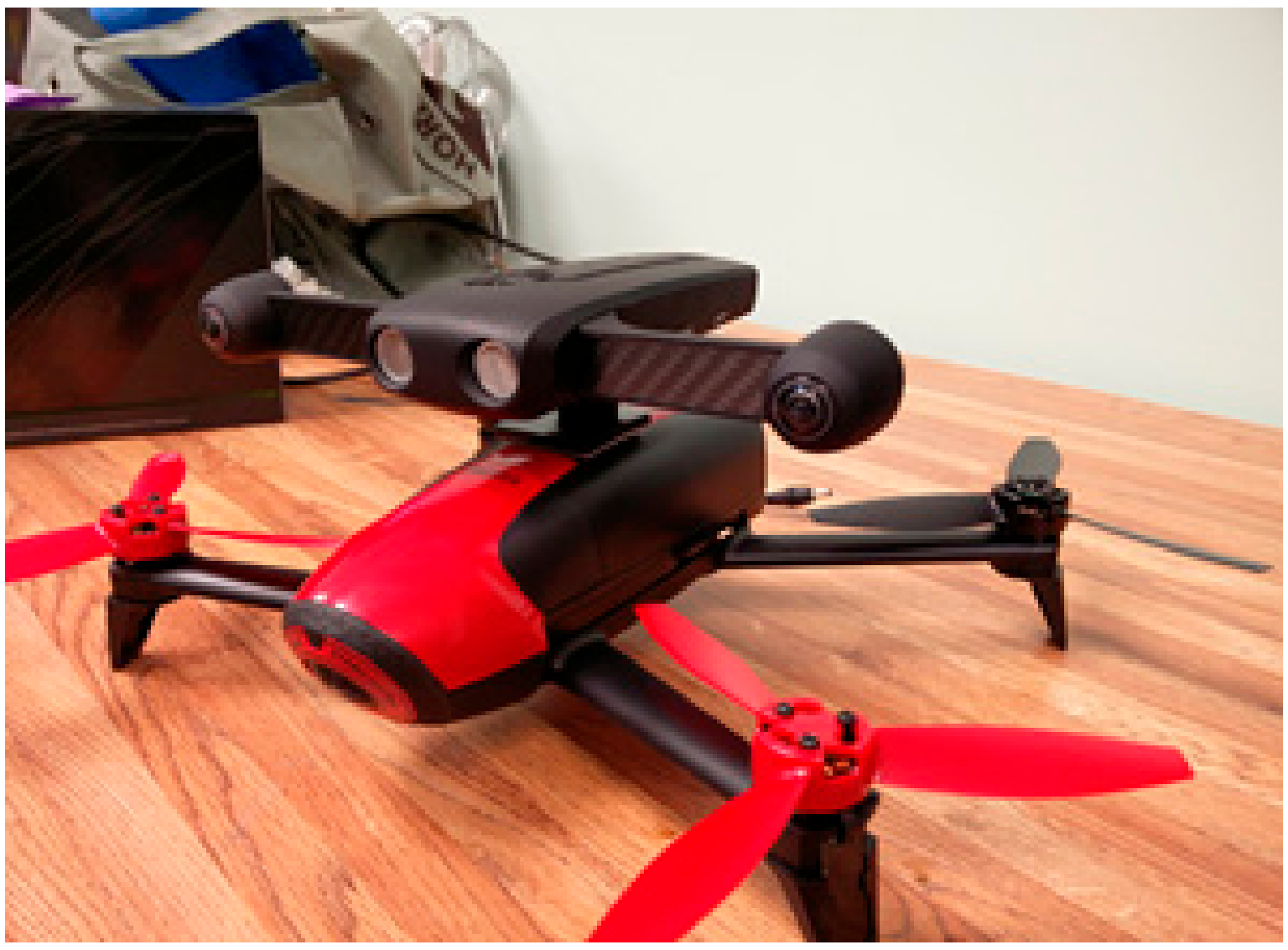

**Figure 8.** The testing hardware setup—A SLAM dunk mounted on a Parrot Bebop 2 drone.

The SLAM dunk module controls the drone by connecting to an access point hosted by the Bebop 2 via USB. On the other hand, the connection between the SLAM dunk and the ground control station is established via Wi-Fi using the Bebop's access point as an interface. This means that all the communication between the SLAM dunk and the ground control station is to be routed via the Parrot's server. This is done to make use of the powerful Wi-Fi hardware present on-board the Bebop 2. The complete network configuration of the setup is illustrated in Figure 9.

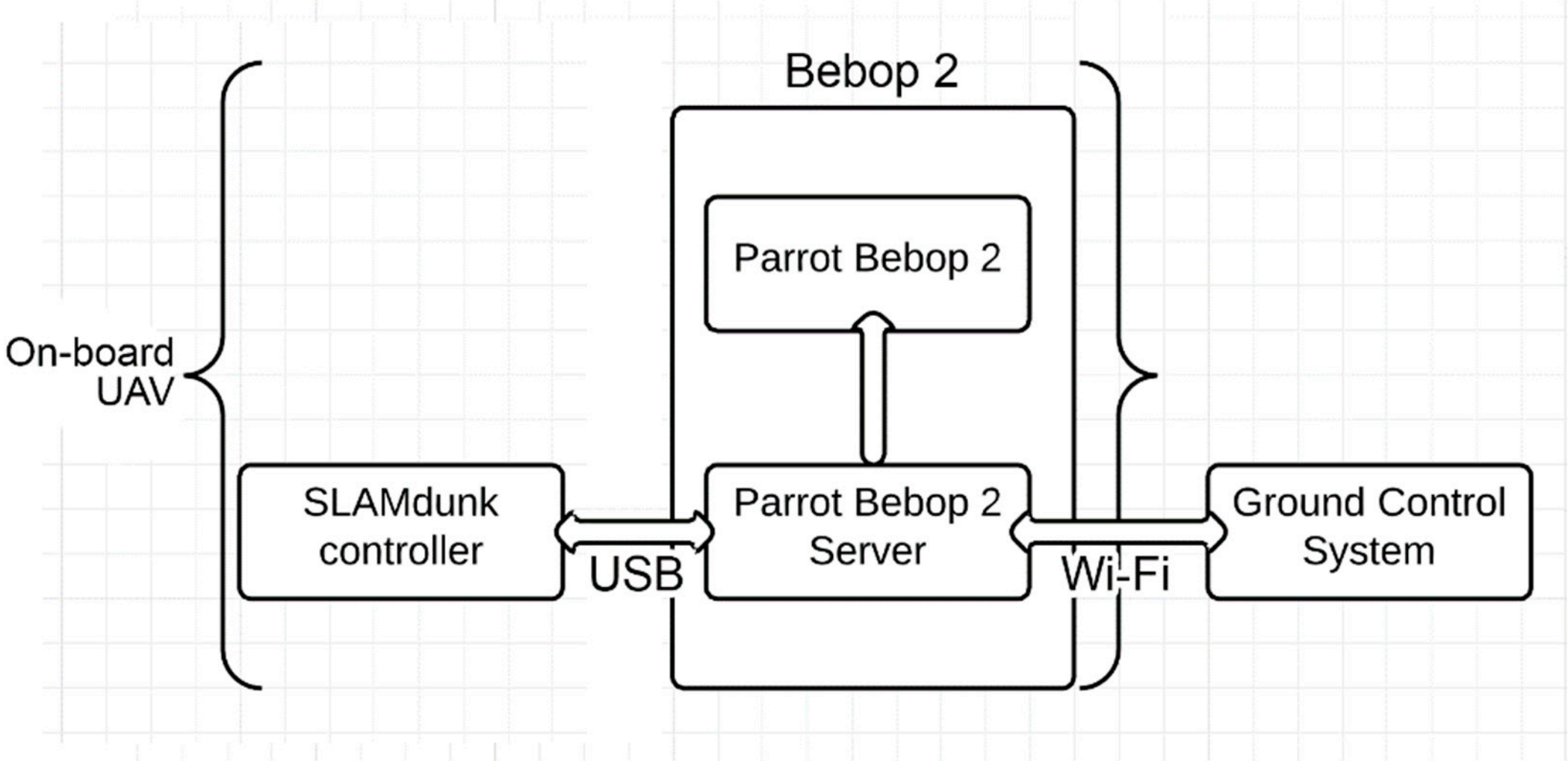

**Figure 9.** Communication setup between Ground Control, SLAM dunk, and Parrot Bebop 2.

The UAV is set to have two main modes of operation: exploration mode and localized mode. In the exploration mode, the UAV's operation is based completely on the SLAM map that is being generated during flight while the localizing sequence continuously attempts to localize the UAV with the CAD model. A flow chart of this mode of operation including the pre-processing steps discussed in Section 2.1 can be seen in Figure 10. Once localized, the UAV switches to localized mode where it is made to follow a set of calculated waypoints (based on its localized position) to a goal location.

### 3.1. Preprocessing

As explained in Section 2.1 for the macro-feature vector matching to occur the BIM of the building needs to be pre-processed. This is done experimentally by loading the BIM into Assimp to be able to access the information within the model. Since Assimp is written in C++ it is easy to integrate into the process pipeline.

Since the BIM format already contains all the macro-features of the environment Figure 1, the next step was to extract the position of the centroids of the macro-features and store them in a k-d Tree which facilitated the creation of the macro-feature description vector for each of the macro-features as discussed explained previously using Algorithm 1. These vectors are later used to find the correspondences between the SLAM and the CAD maps.

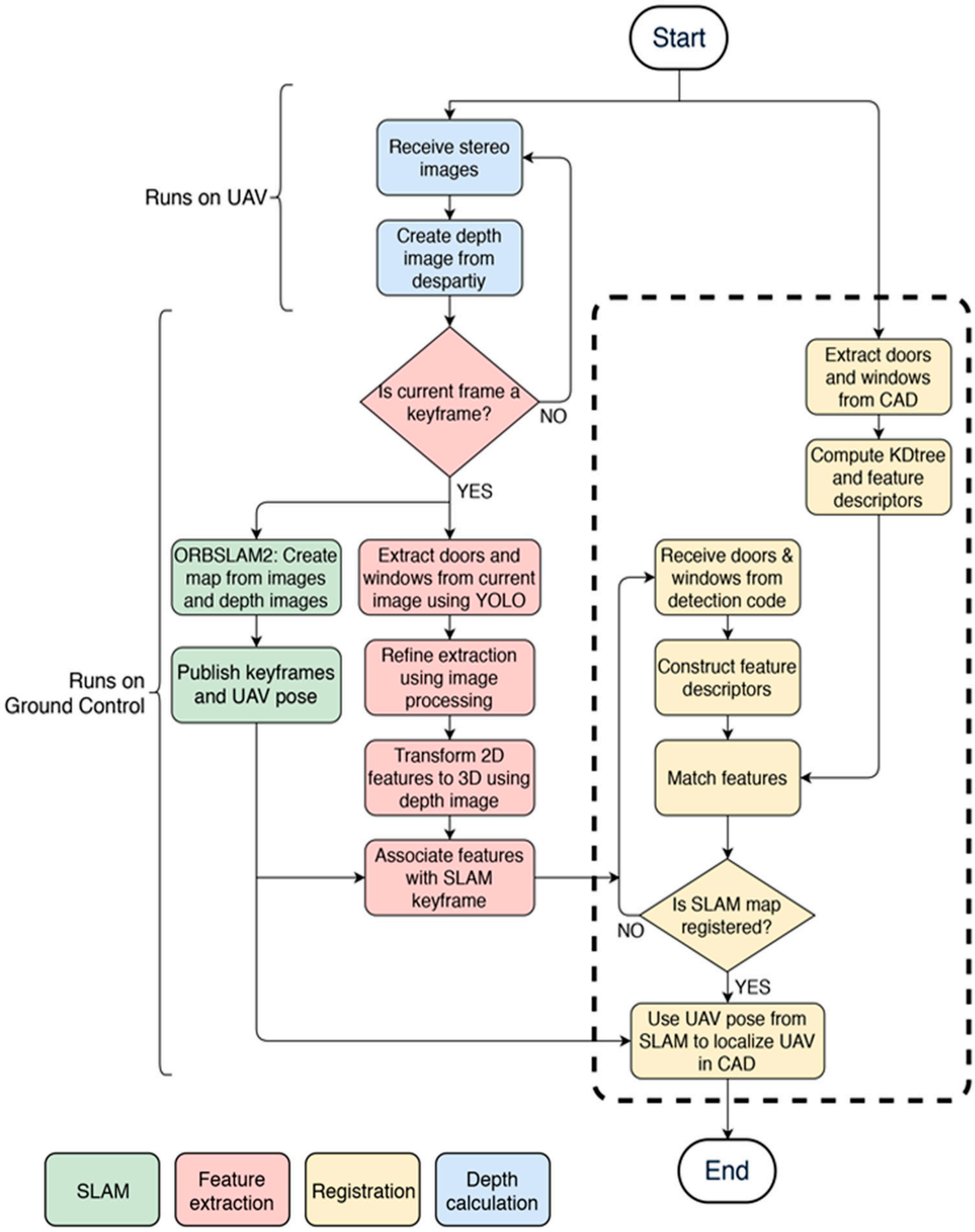

**Figure 10.** Pipeline process flow.

*3.2. Exploration Mode*

In exploration mode, the left and right images from the stereo cameras were used to compute depth from disparity using GPU optimized Semi Global Block Matching (SGBM) [23] on-board the SLAM dunk. The left image and the computed depth images were then sent via Wi-Fi to the ground control station at a rate of 30 fps in addition to IMU updates that were sent over at ~150 Hz. It is to be noted that all communication between the UAV and the ground control including image transfer are handled by Robot Operating System (ROS) [24] communication packets.

On the ground control unit, the system was divided into two main components: the SLAM module and the localization module. The SLAM module estimates the current UAV pose in a map generated using the images and IMU data as illustrated in Figure 11a. ORB-SLAM2 [25,26] was used as the SLAM algorithm and was used in the RGBD mode, which utilized the left and computed depth images from the UAV to construct a map consisting of a sparse set of 3D map points while also estimating the six-degree-of-freedom (6DOF) current UAV pose in it.

The localizer extracted the macro-features discussed earlier (Figure 11b) from the images and tried to localize the UAV in the CAD model by first forming correspondences between the environment's detected macro-features and those extracted from the CAD model. These are then used to estimate a rigid transformation in SE(3) between the two coordinate frames (SLAM and CAD model) using a robust transformation estimation algorithm (Section 2.3.2). The resulting localized UAV is illustrated in Figure 11c.

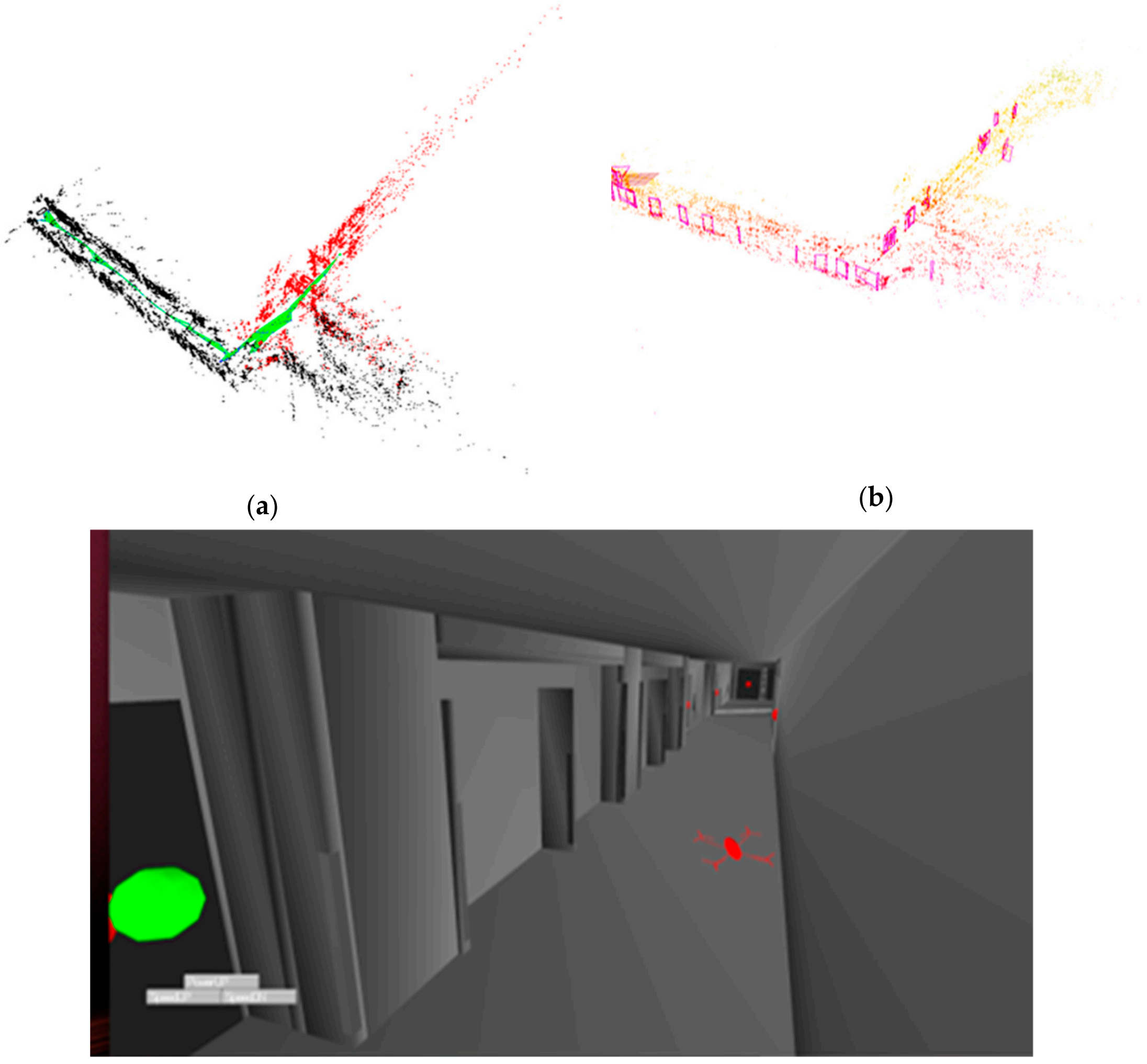

(**a**) (**b**)

(**c**)

**Figure 11.** Exploration mode (**a**) SLAM generated map, (**b**) detected and reconstructed doors in the slam map, and (**c**) localized Unmanned Aerial Vehicles (UAV).

### 3.3. Localized Mode

Once the UAV was localized using the localizer module, the UAV began the process of traveling to the destination. The height of the UAV from the ground level was retrieved using an ultrasound sensor located beneath the Bebop 2 drone. A cross-section of the current floor, roughly at the UAV's flight height was generated from an octomap [27,28] representation of the building. The octomap provided a probabilistic 3D voxel representation of the environment/building that was updated in real-time with objects that did not appear in the original CAD model. The goal position and the current location of the UAV were then projected onto the 2D occupancy grid that is generated from the cross section generated earlier. An A* search algorithm [29] shown in Figure 12, to plan a path from the current location to the goal position was then run on this grid map. This generated a path of shortest route between the current location and the goal location. Dividing the floor into a 2D grid reduces the search space and consequently provided an efficient method to compute the shortest distance. Using the path generated, waypoints were sent to the UAV representing the centroids of the cells in the 2D grid which were connected by the computed path. Since the UAV did not undergo any fast motions, it was sufficient to supply the waypoints as a set of location coordinates with respect to the CAD coordinate system.

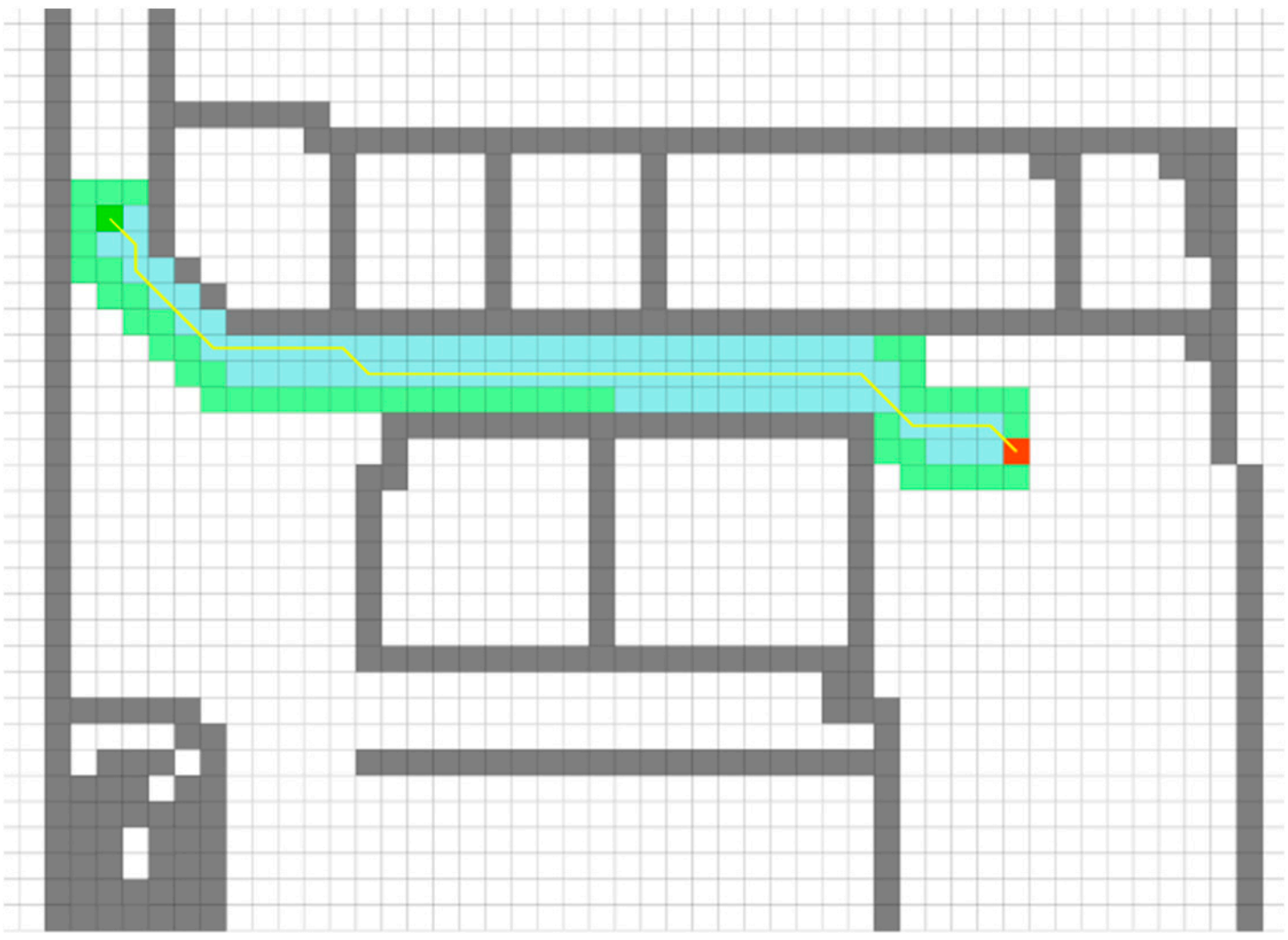

**Figure 12.** A* algorithm for path finding. A cross section of the floor is used to generate the 2D occupancy grid. The dark green cell is the start location and the red cell is the goal location. The gray cells are occupied, and the white cells are free. A path is computed by applying the A* algorithm and waypoints are generated using the centroids of the grid cells.

## 4. Experimental Results and Analysis

The following sections evaluates the performance and accuracy of the different algorithms used. Section 4.1 evaluates the performance of the feature descriptor extraction and matching. Sections 4.2 and 4.3 evaluate the performance of the entire system operation through localization as well as re-localization within the CAD model.

### 4.1. Object Vector Results and Accuracy

A basic method of selecting features based on distances to neighboring features was initially explored. In this method, lookup table of distances from each feature to every other feature was first computed from the CAD model in the preprocessing phase. Two features were then selected at random from the observed dataset and matched with distances computed from the CAD model. On finding a successful match, a new feature was then selected such that the distances to the two selected features was consistent with the distances in the CAD model. This process was repeated until a total of four correspondences were found. This method was found to be resource intensive and slow.

The histogram of the distances to the closest five features from each feature whose descriptor is to be calculated is shown in Figure 13a. The probability density function of the Kernel Density Estimator is shown in Figure 13b. The shape of the KDE closely resembles the histogram and thus represents the underlying data. Dividing the probability density function along the local minima yields 24 different bins for the lookup table.

The accuracy of the matches formed based on the descriptor created from this lookup table has been evaluated across 5 separate runs of the algorithm in different settings and has been shown in Table 1. Each of the matches are formed by finding the feature descriptor with the least Hamming distance from the features in the CAD model.

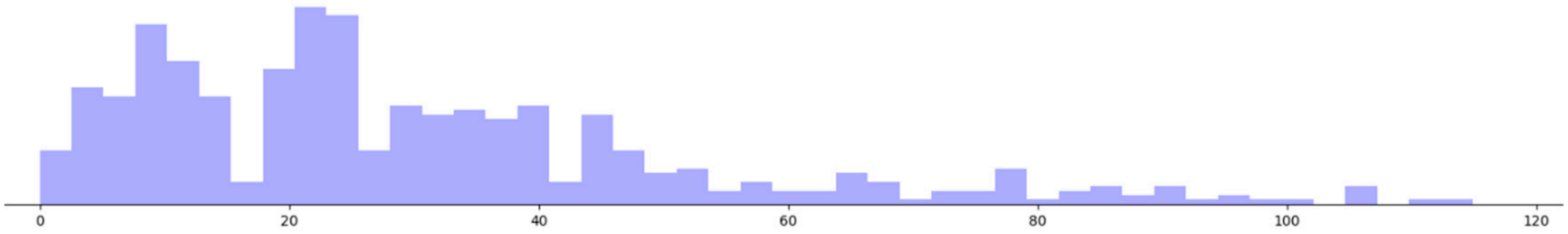

(**a**) The histogram of distances between 5 closest features

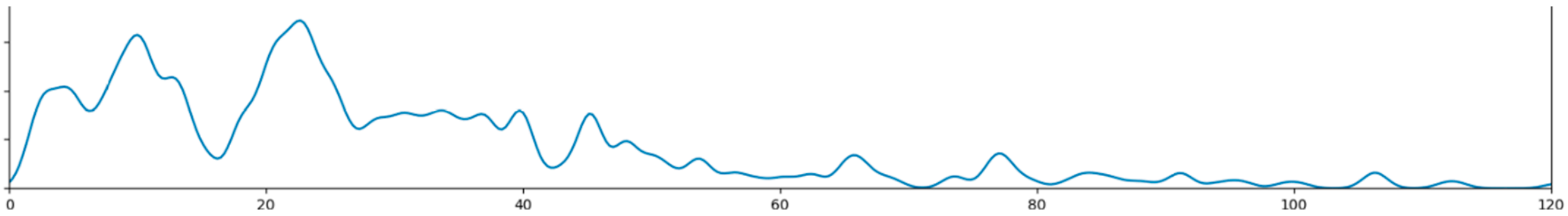

(**b**) The KDE probability density function

**Figure 13.** The histogram and Kernel Density Estimator (KDE) probability density function for distance. The horizontal axis represents the Hamming distance starting from 0. In (**a**), the vertical axis represents the number of features at a particular distance. In (**b**), the vertical axis represents the probability density of a particular distance.

**Table 1.** Feature Descriptor Accuracy.

| Run Id | Features Detected | Correct Matches | Incorrect Matched | Accuracy |
|---|---|---|---|---|
| Run 1 | 21 | 15 | 6 | 71.4% |
| Run 2 | 25 | 19 | 6 | 76% |
| Run 3 | 7 | 4 | 3 | 57.1% |
| Run 4 | 31 | 24 | 7 | 77.4% |
| Run 5 | 19 | 14 | 5 | 73.6% |

It is seen that the matching accuracy increases with the number of features detected. This is expected since increasing the number of features results in a more complete picture of neighboring features which are used to build the feature vector. Lower number of features detected leads to greater chance of ambiguity in the feature vectors due to multiple features having similar distances. Run 3 is an example of such a scenario.

The extraction of the 5 nearest neighbors from the k-d Tree has been found to take an average of approximately 0.052ms on the test system consisting of a 6th generation core-i7 processor. The matching of descriptors using the brute-force matching technique has a complexity of $O(mn)$ and takes approximately 0.3ms for 25 descriptors in the observed dataset and 106 descriptors in the CAD dataset.

### *4.2. Localization within the CAD Model*

The descriptors computed from both the CAD model and the observed data are used to match the two sets of features and calculate the SE(3) rigid transformation between the SLAM map and the CAD model. Figure 14 shows 5 runs of different lengths within the test setup. AprilTags [30,31] were placed at the goal positions for each run. AprilTags are fiducial markers that are not prone to the ambiguities inherent in other markers like Checkerboards or Circlegrids. To measure the accuracy of the localization, the distance between the final location of the UAV and the Apriltag located at the goal position was measured.

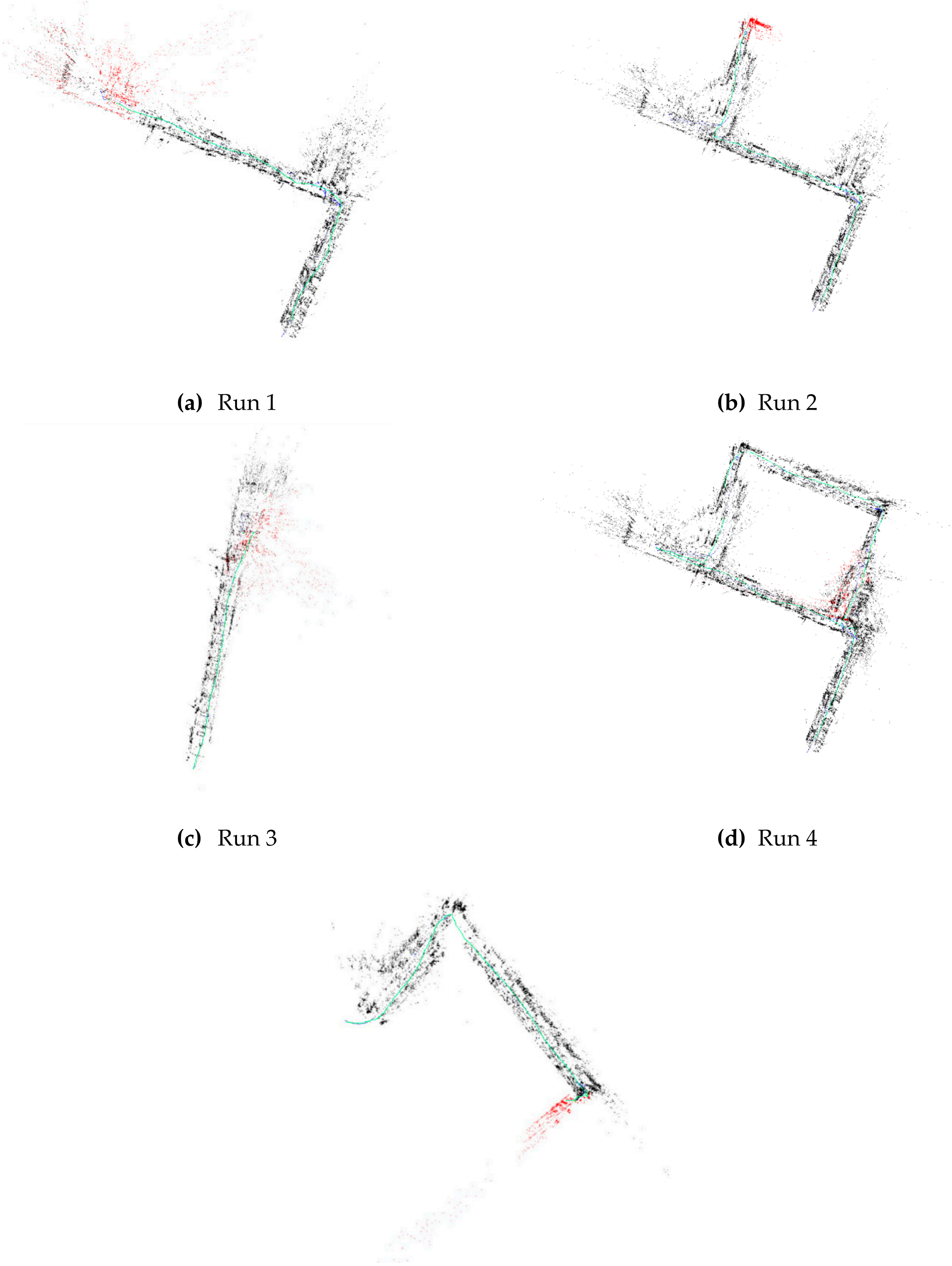

**(a)** Run 1 **(b)** Run 2 **(c)** Run 3 **(d)** Run 4 **(e)** Run 5

**Figure 14.** Runs of different lengths in the test setup. The top-down view of the generated SLAM map is shown here for each of the 5 runs of the system. The run in (d) has been set with multiple goal locations to generate a larger map for use in re-localization testing.

The localization time was calculated based on how long it took the UAV to successfully localize itself in the CAD model from the time the system was started. Table 2 shows the localization time and the error between the system goal position and the actual goal position for each of the runs depicted in Figure 14.

**Table 2.** Localization accuracy.

| Run Id | Localization Time (s) | Error (m) | Error (% of Trajectory Length) |
|---|---|---|---|
| Run 1 | 16 | 0.25 | 4% |
| Run 2 | 17 | 0.14 | 3% |
| Run 3 | 13 | 0.23 | 3% |
| Run 4 | 15 | 0.15 | 1% |
| Run 5 | 13 | 0.26 | 2% |

The UAV was able to successfully localize itself within the CAD model approximately 15 seconds of starting on average. Run 4 had been performed with multiple goal locations to generate the complete map of the floor. This map is then used in a run to test the re-localization mode of the UAV.

*4.3. Re-Localization within the CAD Model*

To test the re-localization capability of the system, a SLAM map was generated from a previous run and its transformation with respect to the CAD model was stored. These were then used when running the UAV a second time. The system was able to quickly localize the UAV within a second of the system initialization. This is possible because of ORB-SLAM2 uses a fast and efficient re-localization module. The initial SLAM map and the view from the perspective of the camera is shown in Figure 15. The real-time experimental results can be viewed in Video S1.

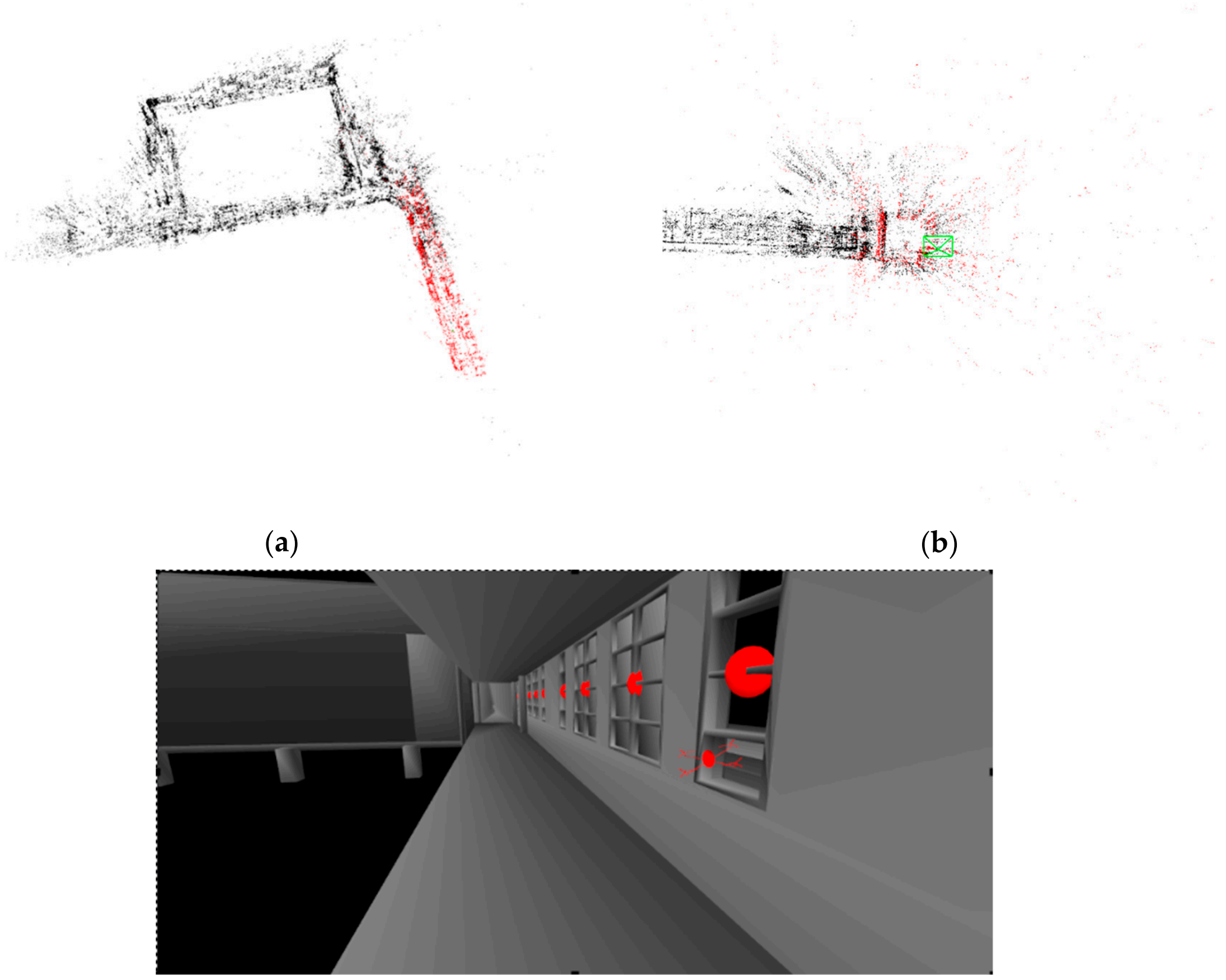

(**a**) (**b**)

(**c**)

**Figure 15.** Re-localization of the UAV in previously constructed map. The previously constructed map is shown in (**a**). The UAV is shown localized in this map in (**b**) while (**c**) shows its representation in 3D CAD model.

## 5. Conclusions

The proposed system utilized a pre-existing scale accurate SLAM system using stereo images to estimate depth, and a pre-existing real-time Convolutional Neural Network for object detection coupled with a novel and computationally efficient method of registration of the SLAM map with the CAD model.

This paper proposed a novel pipeline combining a real-time CNN based object detection network, a SLAM system and a novel registration mechanism combined with image processing techniques to localize a UAV in a CAD model. Furthermore, the paper proposed a computationally efficient, orientation invariant feature descriptor to match features in the CAD model and the observed data, based solely on the spatial correlation between the features.

A major contribution of this work has been the fast and efficient computation of orientation invariant feature descriptors that were used to form correspondences between the observed features and the features in the CAD model. The use of a k-d Tree enabled the quick extraction neighboring features. The use of KDE to group clusters of discrete values together to reduce the time needed to match features had played a crucial part in improving the efficiency and performance of the algorithm.

The experimental validation shows that leveraging readily available 3D CAD models of buildings provided valuable information regarding the environment of the UAV. It provided the system with all the information needed to deduce the current location of the UAV within the CAD model, and enabled the UAV to find its pose without the use of unreliable GPS in indoor environments or other costly systems that emulate the GPS mechanism indoors. The use of a SLAM system registered to a CAD model also enabled the system to leverage the re-localization feature of the SLAM, thus enabling fast localization in previously visited locations.

### *Future Work*

The system suffers from a few limitations that could be improved. The accuracy of the feature descriptor could be improved by applying a threshold on the computed Hamming distance. This could be calculated by saving all the computed descriptors and then studying the result of using different thresholds varying from 0 to 64 for the descriptor matches. All these matches could then be used to generate a Receiver Operating Characteristics (ROC) curve, which could be used to deduce the best possible value for the threshold distance.

The system performance could also be significantly improved by including other macro-features like drinking fountains, exit signs, posts, and other macro-features. Including more macro-features would enable the UAV to localize quickly due to the abundance of macro-features as well as the variations in the macro-feature types which would give rise to less ambiguity in the macro-feature descriptors. This would also alleviate problems arising from symmetries. Furthermore, since the system was developed for a single floor localization problem, added macro-features unique to each floor can further help the localization problem. Similar floors can be a challenging problem with the presented algorithm which is also in consideration in our future work in progress.

**Supplementary Materials:** The following are available online at https://youtu.be/OR3acLQsoBY , Video S1: Autonomous Localization of UAV in a CAD Model.

**Author Contributions:** Conceptualization, A.H. and J.N.; methodology, A.H.; software, A.H.; validation, A.H., writing—original draft preparation, A.H.; writing—review and editing, T.E., A.E., and M.E.; supervision, J.N.

**Funding:** This research was funded by State of North Dakota under Research ND grant.

**Conflicts of Interest:** The Authors declare no conflict of interest.